\documentclass[11pt,twoside]{article}
\usepackage{preprint}
\begin{document}

% Set the page style to empty (no headers or footers)
\thispagestyle{empty}  

%-------------------------------------------------------------------------------
% Title, Author, and Affiliation
%-------------------------------------------------------------------------------

% Create a colored box with specific background and frame colors, and sharp corners
%\begin{tcolorbox}[colback=white!20,colframe=gray!20,sharp corners]

% Add vertical space adjustment
\vspace{-2mm} 
% Create a minipage for the title
\begin{minipage}{15.0cm}
    \begin{center}
        % Title of the document
        \textbf{\large Benchmarking Physics-Informed Neural Networks and Boundary Element Method: Accuracy\allowbreak--\allowbreak Efficiency Trade-offs in Wave Scattering}  
    \end{center}                 
\end{minipage} 
% Create a minipage for the first logo
\begin{minipage}{0.0cm}
    \begin{center}
        % Include the first logo with specific width
        %\hspace{-0.2cm}\includegraphics[width=2.8cm]{figs/format/eafit.pdf}
    \end{center}
\end{minipage}    
 
% Add vertical space adjustment
\vspace{-2mm} 
%\end{tcolorbox}

% Add vertical space adjustment
\vspace{0.6cm}
 
{\footnotesize\noindent O. Rincón-Cardeño$^1$, G. Pérez-Bernal$^1$, S. Montoya-Noguera$^2$, and N. Guarín-Zapata$^{1,*}$ \\
$^1$ Mathematical Applications in Science and Engineering Research Group, School of Applied Sciences and Engineering, Universidad EAFIT, 
  Medellín, Colombia \\
$^2$ Nature and Cities Research Group, School of Applied Sciences and Engineering, Universidad EAFIT, Medellín, Colombia\\
  $^*$ Correspondence: nguarinz@eafit.edu.co}

%-------------------------------------------------------------------------------
% Abstract of the document
%-------------------------------------------------------------------------------

% Create a colored box with specific background and frame colors, and sharp corners
\begin{tcolorbox}[colback=gray!20, colframe=gray!20, sharp corners]

    % Begin the abstract section
    \begin{abstract}

        % Add vertical space adjustment
        \vspace{-1mm}

        % No indentation for the first paragraph
        \noindent
        % Abstract content      
\noindent Physics-informed neural networks (PINNs) offer a flexible alternative to classical numerical solvers for partial differential equations, but their computational trade-offs remain incompletely quantified. We present a controlled benchmark of PINNs and the Boundary Element Method (BEM) for two-dimensional wave scattering problems governed by the Helmholtz equation, applying both approaches to the same problem under identical conditions. BEM solutions are obtained through boundary discretization and system assembly, while PINNs are trained by minimizing residuals of the governing equation and boundary conditions, with network hyperparameters selected through optimization. At comparable accuracy levels, BEM assembly and solution require on the order of $10^{-2}~ \mathrm{s}$, whereas PINNs training incurs costs on the order of $10^{2}~\mathrm{s}$. Once trained, PINNs enable rapid field evaluation at interior points, with inference times comparable to or faster than BEM. These results establish a reproducible benchmarking procedure, providing quantitative guidance for comparing standard and neural network-based methods.\\
 
 \vspace{-1mm}
    \end{abstract}
\end{tcolorbox} 


%-------------------------------------------------------------------------------
% Section - Introduction
%-------------------------------------------------------------------------------
\section*{Introduction}\label{sec:introduction}
\fancyhead[RO]{\textit{Introduction}\, / \thepage} 
%-------------------------------------------------------------------------------

\lettrine[lines=2,loversize=0.15,findent=0.4em,nindent=0em]{\textcolor{gray}{P}}{}artial differential equations (PDEs) are central to mathematical modeling and are used to represent physical processes across a wide range of scientific and engineering disciplines. In particular, wave scattering problems arise in a broad range of fields, including acoustics, electromagnetism, geophysics, and medical imaging. Numerical methods such as the Finite Element Method (FEM) and the Boundary Element Method (BEM) are commonly used to model these problems, as they provide accurate approximations of the solution \citep{bermudez_optimal_2007,chandler-wilde_numerical-asymptotic_2012}. However, their computational cost increases significantly for complex geometries or high-frequency regimes, where fine spatial discretization is required to preserve accuracy.\\ 

These limitations have motivated the exploration of alternative approaches, including machine learning-based methods, which aim to reduce the need for fine discretization and scale more efficiently with the dimensionality of the problem \citep{nganyu_tanyu_deep_2023}. Machine learning has increasingly been applied to the solution of forward problems governed by partial differential equations. Physics-Informed Neural Networks (PINNs) incorporate the governing equations and available data into the training process \citep{lagaris_artificial_1998,dissanayake_neural-network-based_1994,raissi_physics-informed_2019}. PINNs have been used to approximate solutions of PDEs such as the Helmholtz equation \citep{escapil-inchauspe_hyper-parameter_2023,nair_multiple_2024}. However, their performance is often assessed in isolation, without systematic comparison to established numerical methods or using limited, and potentially biased benchmarks. A systematic review in fluid mechanics reports that 79\% of machine learning solvers claiming superior performance over traditional methods do not use adequate baselines \citep{mcgreivy_weak_2024}.\\

Comparisons must be designed to avoid systematically underestimating or overestimating the performance of either approach. For example, this can be achieved by comparing solution times across methods with different accuracies, or by benchmarking against numerical methods that are not optimal for the specific problem. Therefore, the use of a strong, problem-appropriate baseline is essential to assess the true capabilities and limitations of machine learning methods for solving PDEs. For scattering problems governed by the Helmholtz equation, the BEM provides a particularly appropriate reference. In contrast to domain-based methods such as FEM, BEM enforces the radiation condition at infinity and requires discretization only on the boundary of the scatterer. This reduces the dimensionality of the problem and eliminates the need for artificial boundary truncation. For these reasons, BEM is an appropriate baseline for evaluating the performance of PINNs in the case of scattering problems.\\

An important characteristic of machine learning methods such as PINNs is their potential to approximate solutions beyond the stabilized training domain \citep{bonfanti_generalization_2024}. This allows the model to estimate the solution at locations different from the original training points. However, unlike BEM, PINNs do not directly enforce the Sommerfeld radiation condition, which can result in inaccuracies in the far field, such as incorrect decay of the scattered wave at large distances. To address this issue, domain truncation techniques such as Perfectly Matched Layers (PML) or Absorbing Boundary Conditions (ABC) are typically employed \citep{berenger_perfectly_1994,bermudez_optimal_2007,komatitsch_unsplit_2007}. Therefore, a direct comparison between methods is necessary to evaluate their performance and to understand how each approach handles the physical constraints associated with scattering problems.\\

% This context motivates the central research question of this study: How can modern machine learning–based approaches for solving partial differential equations be compared with standard numerical methods? In this work, we focus on the comparison between the Boundary Element Method (BEM) and Physics-Informed Neural Networks (PINNs) for solving acoustic wave scattering problems governed by the Helmholtz equation, which serves as a canonical test case in computational physics. While the numerical experiments are formulated for this specific problem, the proposed comparison framework is methodological and applicable to other classes of partial differential equations and numerical or machine learning–based solvers. Addressing this question requires benchmarks that jointly account for accuracy, computational cost, and generalization behavior. Consistent evaluation criteria enable the identification of the conditions under which each method is better suited to a given problem.\\

Our study is framed within the broader question of: How can machine learning–based approaches for solving partial differential equations be compared with standard numerical methods? In this context, we are particularly interested in addressing the following research question: How do PINNs and the BEM compare in terms of accuracy and computational efficiency for wave scattering problem? In this work, we focus on this case, which serves as a canonical test in computational physics. While the numerical experiments are formulated for this specific problem, we propose a methodological framework for comparison that applies to other classes of partial differential equations and to both numerical and machine learning–based solvers. Addressing this question requires benchmarks that jointly account for accuracy, computational cost, and generalization behavior. Consistent evaluation criteria enable the identification of the conditions under which each method is better suited to a given problem.\\

% This work presents a reproducible benchmarking procedure between BEM and PINNs. The objective of the comparison is to determine the range of conditions where each method performs best and to outline their respective advantages and limitations. Both approaches were applied to the problem of plane wave scattering by an obstacle. The behavior of each method was evaluated both inside and outside the training domain, with emphasis on the asymptotic decay of the scattered field. By explicitly quantifying accuracy–efficiency trade-offs, this study aims to provide practical guidance for the informed adoption of physics-informed machine learning methods alongside classical numerical techniques. 

In this work, we present a reproducible benchmarking procedure between BEM and PINNs. The objective of the comparison is to determine the range of conditions where each method performs best and to outline their respective advantages and limitations. Both approaches were applied to the problem of plane wave scattering by an obstacle. Particularly for PINNs, we adopt the original formulation introduced by  Raissi et al. \cite{raissi_physics-informed_2019}, as a baseline for comparison. We also implemented a hyperparameter optimization strategy to the Helmholtz equation in wave scattering problems. The behavior of each method was evaluated both within and outside the training domain, with an emphasis on the asymptotic decay of the scattered field. With this work, we aim to provide tools for comparative analysis in the field. By explicitly quantifying accuracy-efficiency trade-offs, this study provides practical guidance for the use scientific machine learning methods alongside classical numerical techniques.

%-------------------------------------------------------------------------------
% Section - Wave scattering
%-------------------------------------------------------------------------------
\section{Wave Scattering and the Helmholtz Equation}\label{sec:problem-description}
\fancyhead[RO]{\textit{Wave Scattering and the Helmholtz Equation}\, / \thepage}
%-------------------------------------------------------------------------------

We consider the classical problem of acoustic wave scattering by a single circular obstacle under a time-harmonic incident wave \citep{morse1986theoretical}. The problem involves solving the Helmholtz equation under boundary conditions that represent the interaction between the incident wave and the obstacle. The scattered wave field depends on the geometry of the obstacle and the properties of the surrounding medium, which in this case is unbounded. We illustrate this phenomenon in the case of a long cylinder in \autoref{fig:acoustic-scattering}~(left). In numerical methods based on spatial discretization, the infinite domain is typically approximated by truncating the domain at a finite distance from the obstacle and imposing special boundary conditions to minimize artificial reflections. We begin by considering the Helmholtz equation:

\begin{equation}\label{Helmholtz}
\Delta u(\textbf{x}) + k^{2}u(\textbf{x}) = 0, \hspace{4mm} \textbf{x} \in \Omega_{\text{P}}\, ,
\end{equation}  

\noindent where $\mathbf{x} \in \mathbb{R}^{2}$ is the spatial coordinate, $u(\mathbf{x})$ is the unknown field, $\Delta$ is the Laplacian operator, $k = \frac{\omega}{c}$ is the wavenumber, $\omega$ is the angular frequency, $c$ is the speed of sound in the medium, and $\Omega_{\text{P}}$ is the physical domain where the wave propagates. It is important to note that the Helmholtz equation models a time-harmonic regime. The formulation assumes a steady-state oscillation at a fixed frequency $\omega$, and the entire problem is posed in the frequency domain. \autoref{fig:acoustic-scattering}~(right) shows the representation of the scattering problem considering the domain truncation. The field is obtained by considering the incident and the scattered fields due to the interaction with the obstacle. Therefore, we write it as: 

\begin{figure*}[ht!]
\centering
    \includegraphics[width=6.2 in]{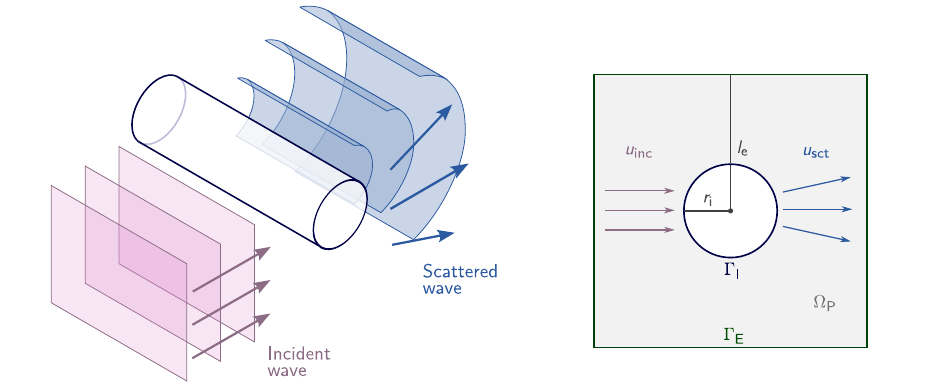}
    \vspace{-1pt}
\caption{
Schematic representations of acoustic wave scattering by an obstacle. 
(left) Three-dimensional illustration of an incident plane wave propagating toward a circular obstacle and generating a scattered field upon interaction with the interface. 
(right) Two-dimensional computational model, where the scattered field $u_{\mathrm{sct}}$ is generated by an incident wave $u_{\mathrm{inc}}$ in the domain $\Omega_{\mathrm{P}}$, bounded internally by a circular interface $\Gamma_{\mathrm{I}}$ of radius $r_{\mathrm{i}}$ and externally by a square boundary $\Gamma_{\mathrm{E}}$ of semi-length $l_{\mathrm{e}}$. 
The scattered field is evaluated within the computational domain $\Omega_{\mathrm{P}}$.
}
\label{fig:acoustic-scattering}
\end{figure*} 

\begin{equation*}
    u(\textbf{x}) = u_{\text{inc}}(\textbf{x})  + u_{\text{sct}}(\textbf{x}) \, .
\end{equation*}
 
The incident wave is \( u_{\text{inc}} \), which is typically a known function describing the incoming acoustic field. In our case, it is a plane wave that we can write in polar coordinates as:

\begin{equation}\label{inc}
u_{\text{inc}}(r,\theta) = A e^{i\mathbf{k} r \cos \theta}\, ,
\end{equation}

\vspace{-5pt}

\noindent where \( A \) is the amplitude and \( k \) is the wavevector. Using the Jacobi-Anger expansion, the variables of the right term can be separated and expressed in Equation \ref{inc} as:

\vspace{-5pt}

\begin{equation}\label{inc-2}
 A e^{i\mathbf{k} r \cos \theta} = A \sum_{n=-\infty}^{\infty} i^n J_{n}(r) e^{in\theta} .
\end{equation}

\vspace{-2pt}

Here, \( J_n \) denotes the Bessel function of the first kind of order \( n \). The scattered wave \( u_{\text{sct}} \) satisfies the Helmholtz equation in the exterior of the obstacle and is subject to a boundary condition at the obstacle surface. The specific form of the boundary condition on the surface (at \( r = r_i \)) depends on the type of obstacle. For a sound-hard obstacle, the appropriate condition is of Neumann type:

\begin{equation}\label{neumann}
\nabla u(\textbf{x}) \cdot \textbf{n} = 0, \hspace{4mm} \textbf{x} \in \Gamma_{\text{I}} .
\end{equation}

In this expression, \( \mathbf{n} \) denotes the outward unit normal to the boundary \( \Gamma_{\text{I}} \). To ensure a well-posed problem in an unbounded domain, the solution must also satisfy the Sommerfeld radiation condition. In the case of a plane problem it reads \citep{schot_eighty_1992}:

\begin{equation}\label{Sommerfeld}
    \lim_{r \to \infty} \left( r^{1/2} \left( \frac{\partial u(r,\theta)}{\partial r} - ik u(r,\theta) \right) \right) = 0,
\end{equation}

 \noindent where \( r \) is the radial distance from the center and \( u(r,\theta) \) is the solution to the Helmholtz equation. This condition is mathematically required to ensure the uniqueness of the solution, and thus the well-posedness of the problem. In the case of domain-based methods, the radiation condition is imposed numerically by truncating the domain with an artificial boundary \( \Gamma_{\text{E}} \) and applying an ABC or PML to approximate outgoing waves. The total acoustic field \( u(\mathbf{x}) \) must satisfy the Helmholtz equation (equation~\eqref{Helmholtz}) in the physical domain \( \Omega_{\text{P}} \), the Neumann boundary condition on the surface of the obstacle \( \Gamma_{\text{I}} \) (equation~\eqref{neumann}), and the Sommerfeld radiation condition as \( r \to \infty \) (equation~\eqref{Sommerfeld}). Together, these three conditions fully define the mathematical formulation of the acoustic scattering problem in an unbounded domain.

%---------------------------------------------------------
\section{Methods}
\label{sec:methods}

This section presents two distinct numerical approaches for solving the exterior Helmholtz problem: the BEM and PINNs. We provide implementation details and discuss numerical strategies for both methods. For PINNs, we detail the procedure used for hyperparameter optimization in order to define a suitable network configuration. Finally, we outline the criteria used to compare the performance of both approaches in terms of accuracy and computational cost. All code used in this study is available in the public repository at \href{https://github.com/oscar-rincon/comparative-bem-pinns}{github.com/oscar-rincon/comparative-bem-pinns}.
 
\subsection*{Boundary Element Method Solution}
\label{sec:methods_bem}

To solve this problem using the BEM, it is necessary to formulate the integral equation for the exterior problem and to establish the boundary conditions for a circular obstacle under a plane incident wave \citep{katsikadelis_boundary_2016, kirkup_boundary_2007}. The following system of integral equations on the boundary is formulated:
 
\begin{equation*}
    \sum_{j=1}^{N} H_{ij}\,u^{j}
    \;=\;
    \sum_{j=1}^{N} G_{ij}\, u_{n}^{j},
\end{equation*}
\noindent where \(u^j\) denotes the boundary solution at node \(j\), and \(u_n^j\) its normal derivative with respect to the outward normal direction. The influence coefficients \(\hat{H}_{ij}\) and \(G_{ij}\) are defined as
\begin{equation*}
\hat{H}_{ij} = \int_{\Gamma_j} \frac{\partial v(p_i, q)}{\partial n_q} \, ds
\end{equation*}

\noindent and

\begin{equation*}
G_{ij} = \int_{\Gamma_j} v(p_i, q) \, ds .
\end{equation*}

\noindent Here, \(p_i\) is the \(i\)-th collocation point on the boundary, and \(q\) is an integration point along the boundary element \(\Gamma_j\). The \(v(p_i,q)\) being the Green’s function for the 2D Helmholtz equation:

    \begin{equation*}
        v  \;=\; \frac{i}{4} H_0^{(1)}(k r),
        \end{equation*}

        \noindent and
        
        \begin{equation*}
        \frac{\partial v}{\partial n_q} \;=\;
        -\frac{i k}{4} H_1^{(1)}(k r)\,\cos\phi.
    \end{equation*}
\noindent where \(r = \|p_i - q\|\) is the distance between the collocation point and the integration point, and \(\phi\) is the angle between the outward normal vector at \(q\) and the vector connecting \(q\) to \(p_i\). The solution at evaluation points is computed as:

\begin{equation*}
\frac{1}{2} u_{\text{pred}}^{i} = \sum_{j=1}^{N} G_{ij} \, u_n^j - \sum_{j=1}^{N} \hat{H}_{ij} \, u^j
\end{equation*}

The influence coefficients are estimated numerically for \( i \ne j \), while for the case \( i = j \), analytical corrections are applied due to the singularity of the kernel. We consider:

\begin{equation*}
\begin{aligned}
G_{jj} = & \left( \int_{\Gamma_j} \frac{i}{4} H_0^{(1)}(k r) ds - \int_{\Gamma_j} \frac{1}{2\pi} \ln{r} ds \right) \\
&+ \int_{\Gamma_j} \frac{1}{2\pi} \ln{r} ds, \quad \text{and} \quad
H_{jj} = 0.
\end{aligned}
\end{equation*}

To approximate the boundary integrals in the computation of the influence coefficients \( G_{ij} \) and \( \hat{H}_{ij} \), we employ Gaussian quadrature over each boundary element \(\Gamma_j\). Each element is mapped to a reference interval, \([-1, 1]\), and the integral is computed as a weighted sum of the integrand evaluated at specific quadrature points:

\begin{equation*}
\int_{\Gamma_j} f(s)\, ds \;\approx\; \sum_{m=1}^{n_q} w_m \, f\big(s(\xi_m)\big) \, |J(\xi_m)|,
\end{equation*}

\noindent where \( \xi_m \) and \( w_m \) are the quadrature points and weights, respectively, and \( J(\xi) \) is the Jacobian of the transformation from the reference element to the physical coordinates. In this work, a Gaussian quadrature rule with \( n_q = 8 \) points is employed for numerical integration.

First, the boundary was discretized into nodes and elements. Then, the solution was computed in the domain of interest, which consists of a uniform grid of evaluation points.

\subsection*{Physics-Informed Neural Networks Solution}
\label{sec:methods_pinns}
 
To solve the boundary value problem, we use PINNs to approximate the scattered field \( u_{\text{sct}} \). Since the domain \( \Omega_{\text{P}} \subset \mathbb{R}^2 \) is unbounded, an artificial boundary \( \Gamma_{\text{E}} \) is introduced to truncate the computational domain. On this boundary, ABC is imposed to approximate the Sommerfeld radiation condition (equation~\eqref{Sommerfeld}):

\begin{equation*}
\nabla u_{\text{sct}}(\mathbf{x}) \cdot \mathbf{n} + ik u_{\text{sct}}(\mathbf{x}) = 0, \quad \mathbf{x} \in \Gamma_{\text{E}}.
\end{equation*}

The full problem is given by the system:

\begin{align}
\Delta u_{\text{sct}}(\mathbf{x}) + k^2 u_{\text{sct}}(\mathbf{x}) &= 0, & \mathbf{x} &\in \Omega_{\text{P}}, \label{problem-1} \\
\nabla u_{\text{sct}}(\mathbf{x}) \cdot \mathbf{n} + \nabla u_{\text{inc}}(\mathbf{x}) \cdot \mathbf{n} &= 0, & \mathbf{x} &\in \Gamma_{\text{I}}, \label{problem-2} \\
\nabla u_{\text{sct}}(\mathbf{x}) \cdot \mathbf{n} + ik u_{\text{sct}}(\mathbf{x}) &= 0, & \mathbf{x} &\in \Gamma_{\text{E}}. \label{problem-3}
\end{align}

Equations~\eqref{problem-1}–\eqref{problem-3} define the physical constraints used to construct the loss function in the PINNs method. A neural network is constructed to approximate the solution 
\( u_{\text{sct}} \colon \mathbb{R}^2 \to \mathbb{C} \), 
using two input variables \((x, y)\), and two output variables corresponding to the real and imaginary parts of the scattered wave: 
\( \Re(u_{\text{sct}}(x,y)) \) and \( \Im(u_{\text{sct}}(x,y)) \). 
The network architecture is a multilayer perceptron implemented in PyTorch \cite{paszke_pytorch_2019}, with weights initialized using the Glorot uniform initializer to promote stable training. 
Training is accelerated on a GeForce RTX 4060 GPU. 
The loss function \( L_{\text{T}} \) is constructed by enforcing the residual of each governing condition at collocation points distributed throughout the domain and its boundaries. In our implementation, all components of the loss function are weighted equally, i.e., no additional weighting coefficients are introduced:

\begin{equation*}
L_{\text{T}} = L_{\Omega_{\text{P}}} + L_{\Gamma_{\text{I}}} + L_{\Gamma_{\text{E}}}.
\end{equation*}
 
The loss function is defined as the sum of squared residuals evaluated at collocation points located in the interior domain \( \Omega_{\text{P}} \), on the obstacle boundary \( \Gamma_{\text{I}} \), and on the artificial boundary \( \Gamma_{\text{E}} \):

\begin{align*}
L_{\Omega_{\text{P}}} &= \frac{1}{N_{\Omega_{\text{P}}}} \sum_{i=1}^{N_{\Omega_{\text{P}}}} \left\Vert \mathcal{R}_{\Omega_{\text{P}}}(\mathbf{x}_{\text{P},i}) \right\Vert^2, \\
L_{\Gamma_{\text{I}}} &= \frac{1}{N_{\Gamma_{\text{I}}}} \sum_{i=1}^{N_{\Gamma_{\text{I}}}} \left\Vert \mathcal{R}_{\Gamma_{\text{I}}}(\mathbf{x}_{\text{I},i}) \right\Vert^2, \\
L_{\Gamma_{\text{E}}} &= \frac{1}{N_{\Gamma_{\text{E}}}} \sum_{i=1}^{N_{\Gamma_{\text{E}}}} \left\Vert \mathcal{R}_{\Gamma_{\text{E}}}(\mathbf{x}_{\text{E},i}) \right\Vert^2,
\end{align*}

\noindent where the residuals are given by:

\begin{align*}
\mathcal{R}_{\Omega_{\text{P}}}(\mathbf{x}) &:= \Delta u_{\text{sct}}(\mathbf{x}) + k^2 u_{\text{sct}}(\mathbf{x}), \\
\mathcal{R}_{\Gamma_{\text{I}}}(\mathbf{x}) &:= \nabla u_{\text{sct}}(\mathbf{x}) \cdot \mathbf{n} + \nabla u_{\text{inc}}(\mathbf{x}) \cdot \mathbf{n}, \\
\mathcal{R}_{\Gamma_{\text{E}}}(\mathbf{x}) &:= \nabla u_{\text{sct}}(\mathbf{x}) \cdot \mathbf{n} + ik u_{\text{sct}}(\mathbf{x}).
\end{align*}

 Here, \( \mathbf{x}_{\text{P},i} \), \( \mathbf{x}_{\text{I},i} \), and \( \mathbf{x}_{\text{E},i} \) denote collocation points in the interior, on the obstacle boundary, and on the artificial boundary, respectively. 
The collocation points are distributed using the Latin hypercube sampling strategy. An analysis of the sampling strategy, including a comparison with an adaptive approach, is provided in Supplementary Material S4 (PINNs sampling points analysis).\\ 

%Each residual enforces the corresponding governing equation or boundary condition in weak form, contributing to the total loss minimized during training. The residual \( \mathcal{R}_{\Omega_{\text{P}}} \) enforces the Helmholtz equation for the scattered field \( u_{\text{sct}} \) in the physical domain \( \Omega_{\text{P}} \), excluding the interior of the obstacle. On the obstacle boundary \( \Gamma_{\text{I}} \), the Neumann condition models a sound-hard obstacle. On the artificial boundary \( \Gamma_{\text{E}} \), the ABC approximates the behavior of an outgoing wave at finite distance and allows truncation of the computational domain without introducing significant reflections. The optimization of the neural network coefficients is performed in two stages: an initial phase using the Adam optimizer to minimize the loss via stochastic gradient descent, followed by a quasi-Newton phase using the L-BFGS algorithm to improve convergence with second-order information.\\ 

\begin{figure*}[ht!]
\centering
    \includegraphics[width=\textwidth]{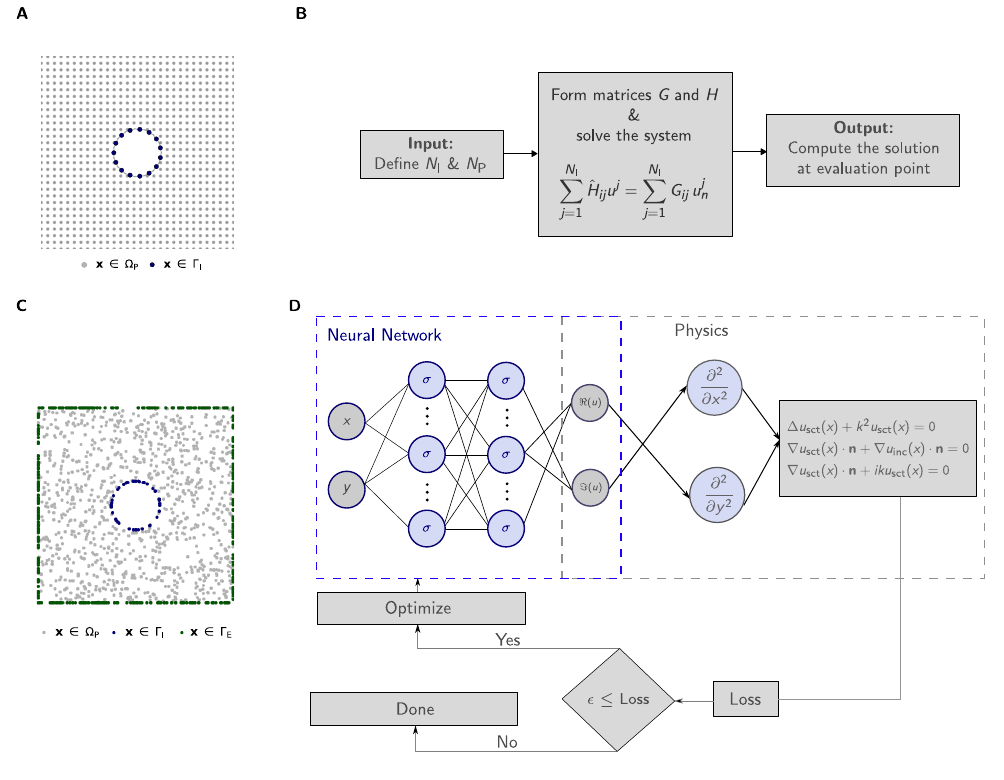}
    \vspace{-12pt} 
\caption{Illustration of the BEM and PINNs implementations for the scattering problem. (A) Distribution of boundary nodes and elements for BEM, along with the spatial arrangement of evaluation points within the domain. Gray dots represent evaluation points, while blue dots indicate integration points on the boundary $\Gamma_I$. (B) Sequential workflow of the BEM implementation. The inputs are the number of integration points for the boundary condition $u_n$ on $\mathbf{x} \in \Gamma_I$ and the number of evaluation points $N_P$. The boundary integral equation is solved to obtain the scattered field, which is then evaluated at the points $\mathbf{x} \in \Omega_P$. (C) Sampling strategy in the physical domain for PINNs, including points in the interior $\Omega_P$ and on the boundaries $\Gamma_I$ and $\Gamma_E$. Gray dots represent collocation points for the Helmholtz equation, while blue dots denote training points for the interior boundary condition on $\Gamma_I$. (D) Optimization process of the PINNs implementation for the Helmholtz equation. The neural network takes spatial coordinates $(x,y)$ as input and outputs the real and imaginary parts of the scattered field $u_{\text{sct}}$. The loss function enforces the physics through the Helmholtz equation in $\Omega_P$, the Neumann boundary condition on $\Gamma_I$, and the Sommerfeld radiation condition on $\Gamma_E$. The optimization loop updates the network parameters until the prescribed tolerance $\epsilon$ is reached.}
    \label{fig:scheme-helmholtz}
\end{figure*}

Both methods differ in their sampling strategies and overall workflows used to estimate the solution. \autoref{fig:scheme-helmholtz} illustrates the implementation of BEM and PINNs for the scattering problem. In the case of BEM, the integration points are uniformly distributed along the interior boundary of the physical domain. The estimation process in BEM occurs in a single computational cycle: it begins with the selection of the number of integration and evaluation points, followed by the assembly and solution of the boundary integral equations. Finally, the solution is evaluated at the specified evaluation points. In contrast, PINNs use randomly distributed training points throughout the physical domain, including the interior, the interior boundary, and the exterior boundary. The estimation process involves multiple optimization cycles, during which the neural network coefficients are iteratively updated to minimize the residuals of the governing equations and boundary conditions.\\

The training process involves several hyperparameters, including the learning rate $\alpha$, 
the number of hidden layers $L$, the number of neurons per layer $N$, and the activation function $\sigma$. These are collectively described by the hyperparameter vector

\[
\lambda = [\alpha, N, L, \sigma],
\]

\noindent where $\lambda \in \Lambda$ and $\Lambda$ denotes the Cartesian product of the predefined hyperparameter ranges. The Adam optimizer relies on gradient updates and is highly sensitive to the choice of learning rate, which directly influences stability and convergence speed during the initial training stage. The network depth $L$ and width $N$ control the representational capacity of the network to approximate more complex functions. However, larger values of $L$ and $N$ increase the number of optimization coefficients, which makes the training process more computationally demanding and can slow down convergence if not properly regularized. Finally, the activation function $\sigma$ affects both the expressiveness of the network and the propagation of gradients, thereby influencing the efficiency of the optimization process. Since it governs the type of nonlinearities introduced at each layer, its choice is particularly important in wave problems such as the Helmholtz equation, where oscillatory behavior must be represented accurately.\\

The ranges for each hyperparameter were selected based on previous related studies in the PINNs literature, ensuring that they encompass configurations known to provide stable and accurate solutions for Helmholtz type problems \citep{escapil-inchauspe_hyper-parameter_2023,nair_multiple_2024}. To determine the optimal configuration, we employed Optuna \cite{akiba_optuna_2019}, an automatic hyperparameter optimization framework compatible with PyTorch. This procedure explored variations in the learning rate, network depth and width, activation functions, and boundary loss weights. 
The final configuration was selected as the one that minimized the objective loss, which is given by the relative error of the prediction. 
 
\subsection*{Comparison strategy}\label{sec:comp-strategy}
 
We conducted numerical experiments to compare the performance of the BEM and PINNs, focusing on solution accuracy and computational cost. We followed a similar approach to that of Grossmann et al. \cite{grossmann_can_2024}. We followed a similar approach to that used by Grossmann et al. (2024)~\cite{grossmann_can_2024}. Both methods were applied to solve the Helmholtz equation and were systematically compared using a ground-truth solution for the scattering obstacle provided in the Section S1 of the Supplementary Material (Analytical Solution).\\

We evaluated the accuracy of BEM as a function of boundary discretization by varying the number of integration points. Smaller element sizes lead to higher accuracy but increase computation time. For PINNs, we controlled model capacity by adjusting the number of layers and neurons per layer, and assessed the corresponding effect on solution accuracy. The expressiveness of a neural network depends on both its depth and width.\\ 

We also estimated the computational time required for both methods. In the case of BEM, this included the time required to assemble and solve the boundary integral system, as well as the time required to evaluate the solution at interior points, while for PINNs, we measured training and evaluation times separately. To compare computational efficiency under similar conditions, we selected the BEM configuration whose accuracy most closely matched that of the best-performing PINN model obtained through hyperparameter tuning.\\

We distinguish between solution time and evaluation time because the BEM and PINNs approximate the solution in different ways. In BEM, the solution is obtained by discretizing the boundary and solving a boundary integral system, while values in the domain are computed afterwards through the numerical evaluation of boundary integral representations. In contrast, in PINNs, a neural network is trained using the collocation points distributed in the domain and on the boundaries. Once the network is trained, the evaluation at new points requires only a forward pass through the network, which is faster than the training phase. For this reason, we report separately the training and evaluation times in PINNs.\\

To analyze the generalization ability of PINNs beyond the training domain and to assess how well each method handles the radiation condition, we perform an additional study. In this analysis, we evaluate the scattered field in a region that extends beyond the training domain by computing the numerical solution at a fixed propagation angle while varying the radial distance from the scatterer. The results obtained from both methods are compared against the analytical solution to assess their accuracy in the far field. We examine the radial decay of the wave amplitude and the relative error as functions of distance, with particular attention given to the role of domain truncation in the PINN framework and its impact on solution accuracy at large distances. 
Based on the results detailed in Section S2 of the Supplementary Material (Hyperparameter Optimization), the configuration $\lambda^\ast = \{\alpha = 10^{-2},\, N = 25,\, L = 3,\, \sigma = \mathrm{Sine}\}$ was selected. 
The hyperparameters were optimized using Optuna over 50 trials, considering the following search ranges: $\alpha \in \{10^{-2}, 10^{-3}, 10^{-4}\}$, $L \in \{1, 2, 3\}$, $N \in \{25, 50, 75\}$, and $\sigma \in \{\mathrm{Tanh}, \mathrm{Sigmoid}, \mathrm{Sine}\}$. 
The trained PINN models used in this study are publicly available in a Zenodo repository \citep{rincon_2026_18351598}.
\\

\section{Results}
\label{sec:results}
\fancyhead[RO]{\textit{Results}\, / \thepage} 

In this section, we present the results obtained from the analytical solution, BEM, and PINNs for the 2D Helmholtz scattering problem. We vary the number of integration points in BEM to achieve comparable accuracy, enabling a fair comparison of computational time. Then, we evaluate the accuracy of both methods outside the training domain to assess their generalization capability. The results highlight the performance of each approach in estimating the scattered field within and beyond the training domain. All experiments were performed on machines equipped with 10 CPU cores / 20 threads (13\textsuperscript{th} Gen Intel Core  i9-13900H @ 3.40 GHz) and an NVIDIA GeForce RTX 4060 Laptop GPU.

\subsection*{Comparison of Accuracy and Computational Efficiency Between Methods}

We conducted a comparative analysis of the BEM and PINNs for solving the two-dimensional Helmholtz equation, focusing on both solution accuracy and computational efficiency under varying parameters. To ensure reproducibility of the accuracy metrics, a fixed random seed was applied during model initialization and training. To account for runtime variability, each experiment was repeated $10$ times and the average execution time was reported.\\
 
We tested BEM with the number of integration points ranging from 5 to 45 in increments of 5, and PINNs with network configurations varying from 1 to 3 hidden layers and 25 to 75 neurons per layer in increments of 25. These ranges are sufficient to capture the error reduction trends and to identify the achievable accuracy scales without making the computation excessively expensive. The remaining hyperparameters of the PINN were fixed according to the optimal configuration identified during the hyperparameter optimization stage. Based on the results detailed in Section S2 of the Supplementary Material (Hyperparameter Optimization), the configuration $\lambda^\ast = [\alpha=10^{-2},\, N=25,\, L=3,\, \sigma=\mathrm{Sine}]$ was selected. The trained PINN models used in this study are publicly available in a Zenodo repository \citep{rincon_2026_18351598}.\\
 
\begin{figure*}[ht!]
    \centering
    \includegraphics[width=\textwidth]{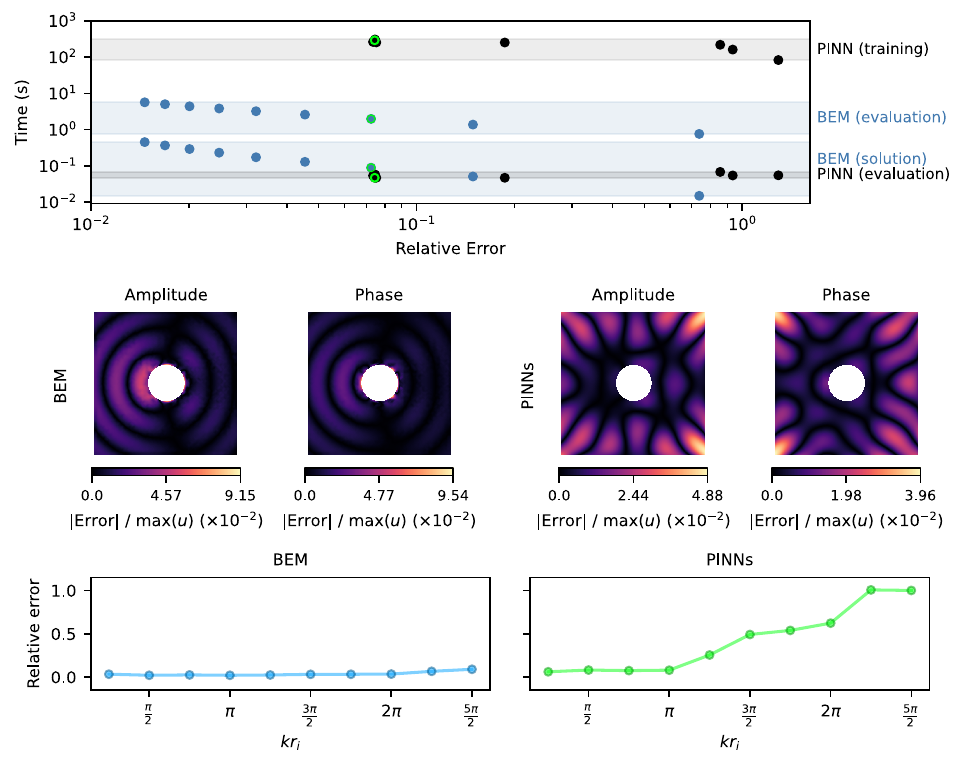}
\caption{Performance evaluation of BEM and PINNs.
(Top) Relative error versus computational time for BEM and PINNs. For BEM, the reported times correspond to the assembly and solution of the boundary integral system and the subsequent evaluation of the solution at interior points. For PINNs, training time and evaluation time are reported separately. Shaded horizontal bands indicate the ranges between the minimum and maximum times for each category. Labels on the right identify the four groups: PINN (training), BEM (evaluation), BEM (solution), and PINN (evaluation). The green highlighted points correspond to the selected configurations, $(L=3, N=25)$ for the PINN and $n=15$ for the BEM, used for direct comparison of computational costs.
(Middle) Spatial distributions of the relative error in amplitude and phase for the selected BEM and PINN configurations.
(Bottom) Relative error as a function of $k r_{i}$ for both methods.}
    \label{fig:error-time}
\end{figure*}

\autoref{fig:error-time}~(top) illustrates the relation between relative error and computational time for BEM and PINNs. As expected, finer discretization in BEM improves accuracy but also increases computational cost. The assembly and solution time for BEM spans from the order of $10^{-1}$ to $10^{-2}$~s across the tested values of $n$, while the evaluation time required to compute the solution at interior points is of $10^{-1}$ to $10^{0}$~s. For PINNs, training times varied between $10^{1}$ and $10^{2}$~s, whereas evaluation times were on the order of $10^{-2}$~s.\\

For the computational time comparison, we selected the BEM configuration that achieved the closest accuracy to the optimal PINN solution. Specifically, the PINN with 3 layers and 25 neurons reached a relative error of $7.44 \times 10^{-2}$, while the closest BEM result was obtained with $n = 15$, yielding $7.25 \times 10^{-2}$. \autoref{fig:error-time}~(bottom) shows the field error for both methods using these configurations. For BEM, the error tends to increase near the interior boundary of the obstacle, whereas for PINNs, errors are mainly concentrated in regions closer to the exterior boundary, where the domain was truncated. We extended the analysis to different frequencies \autoref{fig:error-time}~(bottom). While BEM maintains consistently low errors across the range of $k r_{i}$. PINNs accuracy exhibits a stronger sensitivity to increasing wavenumber.\\

In the selected case, the assembly and solution of the BEM system required a computational time on the order of $10^{-2}$~s, making it approximately four orders of magnitude faster than the training time of the PINN, which was on the order of $10^{2}$~s. This highlights the high computational cost of PINNs associated with the optimization process during training. However, once trained, the PINN model can be evaluated in about $10^{-2}$~s, which is approximately two orders of magnitude faster than the evaluation of the BEM solution at interior points, which requires on the order of $10^{0}$~s. This makes PINNs particularly well suited for applications requiring repeated simulations or near real-time inference. When considering the total computational cost, the PINN requires $289.92$~s (training and evaluation), while the BEM requires $2.06$~s (assembly and evaluation), confirming that the overall cost remains dominated by the PINN training stage. Additional analyses complementing these results are provided in Supplementary Material S3.\\

A central objective of this work was to compare the performance and characteristics of traditional numerical methods and machine learning-based approaches for solving wave scattering problems using a fair baseline. In comparing BEM to PINNs, we addressed two common pitfalls identified in the literature on ML-based PDE solvers \citep{mcgreivy_weak_2024}.\\

First, we ensured that both methods were evaluated at comparable accuracy levels. In the case of the PINN approach, this required an hyperparameter tuning procedure to identify a configuration yielding optimal predictive performance. Once this reference accuracy was established, the number of integration points in the BEM solution was progressively increased until a comparable level of accuracy was achieved. Subsequently, the computational cost of both methods was assessed, thereby satisfying the requirement of comparing methods at either equal accuracy or equal computational cost.\\

Second, to avoid misleading conclusions due to the use of suboptimal numerical baselines, we employed BEM as the reference method, a well-established solver for wave scattering problems. BEM is particularly effective for problems in unbounded domains. In the case they are governed by the Helmholtz equation, it allows the exact enforcement of the Sommerfeld radiation condition through the use of fundamental solutions. Additionally, BEM reduces the dimensionality of the problem by requiring discretization only on the boundary, which is computationally advantageous in unbounded domains. This choice ensures that the traditional method is not only accurate but also representative of state-of-the-art efficiency for the class of problems considered.\\
 
By fulfilling these conditions, our comparative study provides a fair and informative assessment of the trade-offs between both solvers. While BEM demonstrates superior performance in terms of runtime for forward problems, PINNs offer a flexible framework that can be applied to a wide range of scientific and engineering problems, particularly due to the availability of open-source implementations. PINNs employ a mesh-free formulation in which the residuals of the governing equations are minimized at randomly sampled collocation points, allowing flexibility in spatial resolution. In contrast, classical methods such as BEM require mesh refinement to achieve higher accuracy. The proposed PINN framework also presents limitations. In particular, the trained model is problem-specific and depends on the geometry considered during training. Therefore, it cannot be directly evaluated on other geometry without retraining the model. This limitation is further compounded by the fact that retraining PINNs can be computationally expensive due to long training times, as shown in \autoref{fig:error-time}, which restricts their practical applicability. Nonetheless, in scenarios that require repeated evaluations of the same PDE, such as in inverse problems, the relatively fast evaluation time of trained PINNs can be advantageous.\\

It is also important to acknowledge a limitation of this comparison regarding the measurement of computational cost. In this work, efficiency was assessed primarily through wall-clock time, which, although widely adopted in the literature, inherently depends on the characteristics of the hardware and system configuration used to perform the experiments. This makes the results sensitive to differences across machines and may reduce the generalizability of the reported results. A hardware independent approach would be relevant to complement wall-clock time with algorithmic complexity analysis or with estimates of the number of floating-point operations (FLOPs), thereby providing a clearer separation between the intrinsic efficiency of the methods and the capabilities of the underlying hardware. Nevertheless, reporting execution time remains a practical and frequently used benchmark for empirical comparisons, and thus it was employed in this study.\\

The hyperparameter tuning performed presented in the Section S2 of the Supplementary Material (Hyperparameter Optimization) provides further insight into potentially suitable sets of hyperparameters for this problem. However, it is important to note that the performance of PINNs also depends on additional hyperparameters not considered in this work, such as the total number of training iterations and the relative weights of the loss terms. Even with dedicated tuning procedures, the high dimensionality of the design space introduces variability and limits the reproducibility and comparability of results.\\

These factors imply that comparisons between BEM and PINNs should be interpreted in the context of specific problem setups and computational constraints. While BEM yields consistent solutions with relatively low sensitivity to parameters, PINNs require additional effort in model selection and optimization to achieve competitive accuracy. A possible alternative to further improve accuracy would be the incorporation of Fourier features, which have been found appropriate to mitigate spectral bias; however, the purpose of this work was to implement the method as originally proposed. An additional possible approach would be to adopt a multi-objective perspective, taking into account not only accuracy but also computational complexity.\\

\subsection*{Generalization Beyond the Training Domain}\label{sec:appendix-generalization}

\begin{figure*}[ht!]
    \centering
    \includegraphics[width=\textwidth]{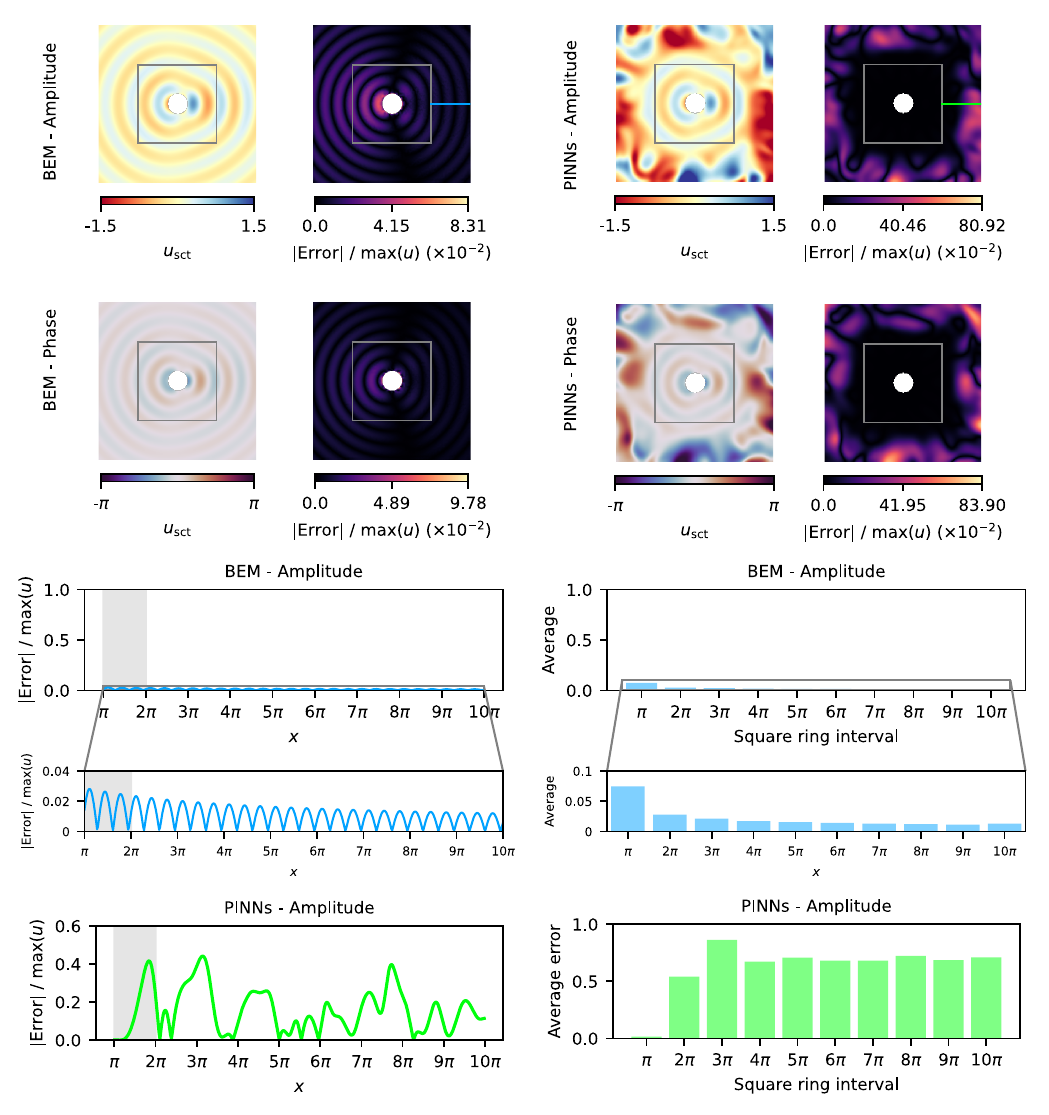}
    \vspace{-8mm}
\caption{
Scattered field computed by BEM and PINNs outside the training region. 
The first two rows show the amplitude (top) and phase (second row) of the scattered field. For each method (BEM on the left, PINNs on the right), the left panel displays the predicted field, while the right panel shows the relative error normalized by the maximum of the reference solution. The square indicates the training domain, corresponding to $l_e = \pi$. 
The third row corresponds to BEM and the fourth row to PINNs. In each case, the left panel shows the relative amplitude error along the line $y = 0$ as a function of $x \in [\pi, 10\pi]$, while the right panel presents the average relative error computed over square ring intervals with semi-lengths ranging from $\pi$ to $10\pi$. For BEM, a zoomed-in view is included to highlight small error variations.
}
\label{fig:general}
\end{figure*}

To evaluate the generalizability of both methods in an extended region of the domain, we computed the solution beyond the original training area. \autoref{fig:general} shows the amplitude and phase of the predicted fields, as well as the spatial distribution of the relative error. For the BEM, the relative error increased to $7.27 \times 10^{-2}$ in the extended region. The error remained uniformly distributed, suggesting that the method maintains consistent accuracy even far from the obstacle. Additionally, an exponential decay of the field amplitude was observed with increasing distance from the scatterer.\\

This behavior reflects the global nature of BEM, which inherently enforces the boundary conditions over the entire domain. In contrast, PINNs experienced a marked reduction in accuracy when evaluated outside the training domain. The relative error increased to 7.09, and the error distribution clearly showed high values in the region outside the training domain.\\

An important distinction between BEM and PINNs lies in how each method handles the radiation condition, which governs the behavior of the scattered field at large distances. In BEM, this condition is satisfied exactly by construction through the use of the fundamental solution in the boundary integral formulation. This allows for an accurate computation of the far-field response without requiring additional assumptions. In contrast, PINNs do not explicitly enforce the radiation condition. Instead, they approximated it using domain truncation techniques, such as the ABC used in this work, which may not fully capture the correct asymptotic behavior of outgoing waves.\\ 

This limitation directly affects the generalization capacity of PINNs beyond the training region. In the current implementation, collocation points are sampled randomly and uniformly within the physical domain. While this strategy is straightforward, it does not prioritize regions where the radiation condition is most critical, particularly near the artificial boundary. As a result, PINNs may not accurately represent the far-field response, especially outside the domain used during training. This lack of generalization is a known limitation of neural network-based models, which approximate solutions only within the region where they are trained. Alternative sampling strategies, such as a generative point sampling strategy \cite{kochliaridis_generative_2025}, or  boundary-focused distributions or adaptive collocation schemes, could help mitigate this issue and improve the enforcement of far-field behavior.\\

%-------------------------------------------------------------------------------
% Section - Conclusions
%-------------------------------------------------------------------------------
\section*{Conclusions}\label{sec:conclusions}
\fancyhead[RO]{\textit{Conclusions}\, / \thepage}
%-------------------------------------------------------------------------------

In summary, while BEM ensures accuracy through the direct satisfaction of physical conditions and dimensional reduction, PINNs offer flexible implementation through the configuration of their loss function, but require careful hyperparameter tuning and strategic sampling to achieve reliable results. Although BEM outperforms PINNs when considering solution time and accuracy, the strengths of each method are complementary, motivating the exploration of hybrid approaches that combine the robustness of boundary-based formulations with the adaptability of neural networks.\\

Recent advances in computational modeling have integrated neural networks as function approximators, supported by automatic differentiation and GPU acceleration. These machine learning techniques offer advantages such as mesh-free implementation, efficient inference, and flexibility to incorporate data into the solution process.\\

Our comparative analysis demonstrates that PINNs require significantly longer training times than BEM to achieve comparable accuracy. This computational cost limits their practical use for forward simulations. However, hybrid approaches that combine the strengths of both frameworks may offer improved performance, particularly in inverse problems, where repeated model evaluations and parameter estimation are required.\\

In future work, we aim to investigate hybrid modeling methods that combine the physical accuracy of BEM with the flexibility of neural networks, building on recent developments in boundary-based and physics-informed approaches \cite{nagy-huber_physics-informed_2024,qu_boundary_2024,li_boundary-based_2025}. Specifically, we propose to replace the differential operator in the PINN loss function with a boundary integral formulation. For the Helmholtz equation, this approach enables the direct incorporation of the Sommerfeld radiation condition into the training process, improving the model’s ability to generalize beyond the training domain and ensuring accurate predictions in unbounded regions. Moreover, this formulation is particularly advantageous in inverse problems, where enforcing correct far-field behavior is essential for achieving stable and physically consistent reconstructions.
 
%-------------------------------------------------------------------------------
% Import references from the bibliography file
%-------------------------------------------------------------------------------
\bibliography{refs}

@article{nair_multiple_2024,
	title = {Multiple scattering simulation via physics-informed neural networks},
	issn = {1435-5663},
	url = {https://doi.org/10.1007/s00366-024-02038-3},
	doi = {10.1007/s00366-024-02038-3},
	language = {en},
	urldate = {2024-10-21},
	journal = {Engineering with Computers},
	author = {Nair, Siddharth and Walsh, Timothy F. and Pickrell, Greg and Semperlotti, Fabio},
	month = jul,
	year = {2024},
}

@article{bermudez_optimal_2007,
	title = {An optimal perfectly matched layer with unbounded absorbing function for time-harmonic acoustic scattering problems},
	volume = {223},
	issn = {0021-9991},
	url = {https://www.sciencedirect.com/science/article/pii/S0021999106004487},
	doi = {10.1016/j.jcp.2006.09.018},
	number = {2},
	urldate = {2024-10-21},
	journal = {Journal of Computational Physics},
	author = {Bermúdez, A. and Hervella-Nieto, L. and Prieto, A. and Rodrı´guez, R.},
	month = may,
	year = {2007},
	pages = {469--488},
}

@book{kirkup_boundary_2007,
	title = {The {Boundary} {Element} {Method} in {Acoustics}},
	volume = {8},
	isbn = {978-0-9534031-0-3},
	author = {Kirkup, Stephen},
publisher = {Integrated Sound Software},
	month = jan,
	year = {2007},
	note = {Journal Abbreviation: Journal of Computational Acoustics
Publication Title: Journal of Computational Acoustics},
}

@book{katsikadelis_boundary_2016,
	address = {London},
	edition = {2nd edition},
	title = {The {Boundary} {Element} {Method} for {Engineers} and {Scientists}, {Second} {Edition}: {Theory} and {Applications}},
	isbn = {978-0-12-804493-3},
	shorttitle = {The {Boundary} {Element} {Method} for {Engineers} and {Scientists}, {Second} {Edition}},
	language = {English},
	publisher = {Academic Press},
	author = {Katsikadelis, John T.},
	month = jul,
	year = {2016},
}

@article{nagy-huber_physics-informed_2024,
	title = {Physics-informed boundary integral networks ({PIBI}-{Nets}): {A} data-driven approach for solving partial differential equations},
	volume = {81},
	issn = {1877-7503},
	shorttitle = {Physics-informed boundary integral networks ({PIBI}-{Nets})},
	url = {https://www.sciencedirect.com/science/article/pii/S1877750324001480},
	doi = {10.1016/j.jocs.2024.102355},
	urldate = {2025-05-20},
	journal = {Journal of Computational Science},
	author = {Nagy-Huber, Monika and Roth, Volker},
	month = sep,
	year = {2024},
	pages = {102355},
}

@article{qu_boundary_2024,
	title = {Boundary integrated neural networks and code for acoustic radiation and scattering},
	volume = {4},
	copyright = {© 2024 The Authors. International Journal of Mechanical System Dynamics published by John Wiley \& Sons Australia, Ltd on behalf of Nanjing University of Science and Technology.},
	issn = {2767-1402},
	url = {https://onlinelibrary.wiley.com/doi/abs/10.1002/msd2.12109},
	doi = {10.1002/msd2.12109},
	language = {en},
	number = {2},
	urldate = {2025-05-19},
	journal = {International Journal of Mechanical System Dynamics},
	author = {Qu, Wenzhen and Gu, Yan and Zhao, Shengdong and Wang, Fajie and Lin, Ji},
	year = {2024},
	note = {\_eprint: https://onlinelibrary.wiley.com/doi/pdf/10.1002/msd2.12109},
	pages = {131--141},
}

@article{raissi_physics-informed_2019,
	title = {Physics-informed neural networks: {A} deep learning framework for solving forward and inverse problems involving nonlinear partial differential equations},
	volume = {378},
	doi = {10.1016/J.JCP.2018.10.045},
	journal = {Journal of Computational Physics},
	author = {Raissi, M. and Perdikaris, P. and Karniadakis, G. E.},
	month = feb,
	year = {2019},
	note = {Publisher: Academic Press},
	pages = {686--707},
}

@article{grossmann_can_2024,
	title = {Can physics-informed neural networks beat the finite element method?},
	volume = {89},
	issn = {0272-4960},
	url = {https://doi.org/10.1093/imamat/hxae011},
	doi = {10.1093/imamat/hxae011},
	number = {1},
	urldate = {2025-06-26},
	journal = {IMA Journal of Applied Mathematics},
	author = {Grossmann, Tamara G and Komorowska, Urszula Julia and Latz, Jonas and Schönlieb, Carola-Bibiane},
	month = jan,
	year = {2024},
	pages = {143--174},
}

@article{escapil-inchauspe_hyper-parameter_2023,
	title = {Hyper-parameter tuning of physics-informed neural networks: {Application} to {Helmholtz} problems},
	volume = {561},
	issn = {0925-2312},
	shorttitle = {Hyper-parameter tuning of physics-informed neural networks},
	url = {https://www.sciencedirect.com/science/article/pii/S0925231223009499},
	doi = {10.1016/j.neucom.2023.126826},
	urldate = {2025-06-26},
	journal = {Neurocomputing},
	author = {Escapil-Inchauspé, Paul and Ruz, Gonzalo A.},
	month = dec,
	year = {2023},
	pages = {126826},
}

@article{chandler-wilde_numerical-asymptotic_2012,
	title = {Numerical-asymptotic boundary integral methods in high-frequency acoustic scattering},
	volume = {21},
	issn = {1474-0508, 0962-4929},
	url = {https://www.cambridge.org/core/journals/acta-numerica/article/abs/numericalasymptotic-boundary-integral-methods-in-highfrequency-acoustic-scattering/5FDA12EF93DD25E89AF7B57A01F8CAEE},
	doi = {10.1017/S0962492912000037},
	language = {en},
	urldate = {2025-06-27},
	journal = {Acta Numerica},
	author = {Chandler-Wilde, Simon N. and Graham, Ivan G. and Langdon, Stephen and Spence, Euan A.},
	month = may,
	year = {2012},
	pages = {89--305},
}

@article{nganyu_tanyu_deep_2023,
	title = {Deep learning methods for partial differential equations and related parameter identification problems},
	volume = {39},
	issn = {0266-5611},
	url = {https://dx.doi.org/10.1088/1361-6420/ace9d4},
	doi = {10.1088/1361-6420/ace9d4},
	language = {en},
	number = {10},
	urldate = {2025-06-27},
	journal = {Inverse Problems},
	author = {Nganyu Tanyu, Derick and Ning, Jianfeng and Freudenberg, Tom and Heilenkötter, Nick and Rademacher, Andreas and Iben, Uwe and Maass, Peter},
	month = aug,
	year = {2023},
	note = {Publisher: IOP Publishing},
	pages = {103001},

}

@misc{mcgreivy_weak_2024,
	title = {Weak baselines and reporting biases lead to overoptimism in machine learning for fluid-related partial differential equations},
	url = {http://arxiv.org/abs/2407.07218},
	doi = {10.48550/arXiv.2407.07218},
	urldate = {2024-07-16},
	publisher = {arXiv},
	author = {McGreivy, Nick and Hakim, Ammar},
	month = jul,
	year = {2024},
	note = {arXiv:2407.07218 [physics]},
}

@article{bonfanti_generalization_2024,
	title = {On the generalization of {PINNs} outside the training domain and the hyperparameters influencing it},
	volume = {36},
	issn = {1433-3058},
	url = {https://doi.org/10.1007/s00521-024-10178-2},
	doi = {10.1007/s00521-024-10178-2},
	language = {en},
	number = {36},
	urldate = {2025-06-27},
	journal = {Neural Computing and Applications},
	author = {Bonfanti, Andrea and Santana, Roberto and Ellero, Marco and Gholami, Babak},
	month = dec,
	year = {2024},
	pages = {22677--22696},
}

@article{lagaris_artificial_1998,
	title = {Artificial neural networks for solving ordinary and partial differential equations},
	volume = {9},
	issn = {1941-0093},
	url = {https://ieeexplore.ieee.org/document/712178},
	doi = {10.1109/72.712178},
	number = {5},
	urldate = {2024-02-06},
	journal = {IEEE Transactions on Neural Networks},
	author = {Lagaris, I.E. and Likas, A. and Fotiadis, D.I.},
	month = sep,
	year = {1998},
	note = {Conference Name: IEEE Transactions on Neural Networks},
	pages = {987--1000},
}

@article{dissanayake_neural-network-based_1994,
	title = {Neural-network-based approximations for solving partial differential equations},
	volume = {10},
	copyright = {Copyright © 1994 John Wiley \& Sons, Ltd},
	issn = {1099-0887},
	url = {https://onlinelibrary.wiley.com/doi/abs/10.1002/cnm.1640100303},
	doi = {10.1002/cnm.1640100303},
	language = {en},
	number = {3},
	urldate = {2025-03-28},
	journal = {Communications in Numerical Methods in Engineering},
	author = {Dissanayake, M. W. M. G. and Phan-Thien, N.},
	year = {1994},
	note = {\_eprint: https://onlinelibrary.wiley.com/doi/pdf/10.1002/cnm.1640100303},
	pages = {195--201},
}

@article{komatitsch_unsplit_2007,
	title = {An unsplit convolutional {Perfectly} {Matched} {Layer} improved at grazing incidence for the seismic wave equation},
	volume = {72},
	doi = {10.1190/1.2757586},
	journal = {Geophysics},
	author = {Komatitsch, Dimitri and Martin, Roland},
	month = sep,
	year = {2007},
}

@article{berenger_perfectly_1994,
	title = {A perfectly matched layer for the absorption of electromagnetic waves},
	volume = {114},
	issn = {0021-9991},
	url = {https://www.sciencedirect.com/science/article/pii/S0021999184711594},
	doi = {10.1006/jcph.1994.1159},
	number = {2},
	urldate = {2025-08-20},
	journal = {Journal of Computational Physics},
	author = {Berenger, Jean-Pierre},
	month = oct,
	year = {1994},
	pages = {185--200},
}

@article{schot_eighty_1992,
	title = {Eighty years of {Sommerfeld}'s radiation condition},
	volume = {19},
	issn = {0315-0860},
	url = {https://www.sciencedirect.com/science/article/pii/031508609290004U},
	doi = {10.1016/0315-0860(92)90004-U},
	number = {4},
	urldate = {2025-08-20},
	journal = {Historia Mathematica},
	author = {Schot, Steven H},
	month = nov,
	year = {1992},
	pages = {385--401},
}

@book{morse1986theoretical,
  title = {Theoretical Acoustics},
  author = {Morse, Philip McCord and Ingard, K Uno},
  year = {1987},
  publisher = {Princeton University Press},
  isbn = {978-0-691-02401-1},
  langid = {english}
}

@article{kochliaridis_generative_2025,
	title = {Generative point sampling strategies for physics-informed neural networks},
	volume = {41},
	issn = {1435-5663},
	url = {https://doi.org/10.1007/s00366-025-02158-4},
	doi = {10.1007/s00366-025-02158-4},
	language = {en},
	number = {5},
	urldate = {2025-12-11},
	journal = {Engineering with Computers},
	author = {Kochliaridis, Vasileios and Dilmperis, Ioannis and Palaskos, Achilleas and Vlahavas, Ioannis},
	month = oct,
	year = {2025},
	pages = {3219--3239},
}

@article{li_boundary-based_2025,
	title = {A boundary-based fourier neural operator ({B}-{FNO}) method for efficient parametric acoustic wave analysis},
	volume = {41},
	issn = {1435-5663},
	url = {https://doi.org/10.1007/s00366-024-02103-x},
	doi = {10.1007/s00366-024-02103-x},
	language = {en},
	number = {4},
	urldate = {2025-12-11},
	journal = {Engineering with Computers},
	author = {Li, Ruoyan and Ye, Wenjing and Liu, Yijun},
	month = aug,
	year = {2025},
	pages = {2393--2410},
}

@misc{akiba_optuna_2019,
	title = {Optuna: {A} {Next}-generation {Hyperparameter} {Optimization} {Framework}},
	shorttitle = {Optuna},
	url = {http://arxiv.org/abs/1907.10902},
	doi = {10.48550/arXiv.1907.10902},
	urldate = {2025-12-15},
	publisher = {arXiv},
	author = {Akiba, Takuya and Sano, Shotaro and Yanase, Toshihiko and Ohta, Takeru and Koyama, Masanori},
	month = jul,
	year = {2019},
	note = {arXiv:1907.10902 [cs]},
}

@inproceedings{paszke_pytorch_2019,
	title = {{PyTorch}: {An} {Imperative} {Style}, {High}-{Performance} {Deep} {Learning} {Library}},
	volume = {32},
	shorttitle = {{PyTorch}},
	url = {https://proceedings.neurips.cc/paper/2019/hash/bdbca288fee7f92f2bfa9f7012727740-Abstract.html},
	urldate = {2025-12-15},
	booktitle = {Advances in {Neural} {Information} {Processing} {Systems}},
	publisher = {Curran Associates, Inc.},
	author = {Paszke, Adam and Gross, Sam and Massa, Francisco and Lerer, Adam and Bradbury, James and Chanan, Gregory and Killeen, Trevor and Lin, Zeming and Gimelshein, Natalia and Antiga, Luca and Desmaison, Alban and Kopf, Andreas and Yang, Edward and DeVito, Zachary and Raison, Martin and Tejani, Alykhan and Chilamkurthy, Sasank and Steiner, Benoit and Fang, Lu and Bai, Junjie and Chintala, Soumith},
	year = {2019},
}

@misc{rincon_2026_18351598,
  author       = {Rincón, Oscar A. and
                  Pérez-Bernal, Gregorio and
                  Montoya-Noguera, Silvana and
                  Guarín-Zapata, Nicolás},
  title        = {Trained Physics-Informed Neural Network (PINN)
                   Models for the 2D Helmholtz Equation
                  },
  month        = jan,
  year         = 2026,
  publisher    = {Zenodo},
  doi          = {10.5281/zenodo.18351598},
}

\end{document}